\documentclass[conference,twoside]{IEEEtran}
\IEEEoverridecommandlockouts
\usepackage[a4paper,top=0.825in,bottom=0.825in,right=0.68in,left=0.68in]{geometry}
\usepackage{fancyhdr}
\usepackage{epsfig}
\usepackage{graphicx}
\usepackage{amsmath}
\usepackage{amssymb}
\usepackage{hyperref}
\usepackage{cite}
\usepackage{longtable}
\usepackage{tabularx}
\usepackage{array}
\usepackage{caption}
\usepackage{subcaption}
\usepackage{times}
\usepackage[utf8]{inputenc}
\ifCLASSINFOpdf
\else
\fi
\begin{document}
%
\title{Large-scale Datasets: Faces with Partial Occlusions and Pose Variations in the Wild}


\author{\IEEEauthorblockN{Tarik Alafif\IEEEauthorrefmark{1},
Zeyad Hailat\IEEEauthorrefmark{2}, Melih Aslan\IEEEauthorrefmark{3} and
Xuewen Chen\IEEEauthorrefmark{4}}
\IEEEauthorblockA{Computer Science Department,
Wayne State University\\
Detroit, MI, USA 48120\\
Email: \IEEEauthorrefmark{1}tarik\_alafif@wayne.edu,
\IEEEauthorrefmark{2}zmhailat@wayne.edu,
\IEEEauthorrefmark{3}melih.aslan@wayne.edu,
\IEEEauthorrefmark{4}xuewen.chen@wayne.edu}\vspace{1.5ex}}

%


\maketitle

\begin{abstract}
Face detection methods have relied on face datasets for training. However, existing face datasets tend to be in small scales for face learning in both constrained and unconstrained environments. In this paper, we first introduce our large-scale image datasets, Large-scale Labeled Face (LSLF) and noisy Large-scale Labeled Non-face (LSLNF). Our LSLF dataset consists of a large number of unconstrained multi-view and partially occluded faces. The faces have many variations in color and grayscale, image quality, image resolution, image illumination, image background, image illusion, human face, cartoon face, facial expression, light and severe partial facial occlusion, make up, gender, age, and race. Many of these faces are partially occluded with accessories such as tattoos, hats, glasses, sunglasses, hands, hair, beards, scarves, microphones, or other objects or persons. The LSLF dataset is currently the largest labeled face image dataset in the literature in terms of the number of labeled images and the number of individuals compared to other existing labeled face image datasets. Second, we introduce our CrowedFaces and CrowedNonFaces image datasets. The crowedFaces and CrowedNonFaces datasets include faces and non-faces images from crowed scenes. These datasets essentially aim for researchers to provide a large number of training examples with many variations for large scale face learning and face recognition tasks.
\end{abstract}


%
\IEEEpeerreviewmaketitle

\section{Introduction}
%
%
%
%

Face detection methods have relied on face datasets for training. However, existing face image datasets tend to be in small scales for face learning in both constrained and unconstrained environments. In order to obtain magnificent performance in face detection systems, it require a large number of unconstrained face and non-face training examples with many variations from real life scenes for training. Unconstrained face training examples should have variations in color and grayscale, quality, resolution, illumination, background, illusion, human face, cartoon face, facial expression, light and severe partial facial occlusion, make up, gender, age, and race. Hence, we introduce our large-scale image datasets for faces and non-faces extracted from the wild in the unconstrained environments. 

The proposed datasets are used for training LSDL face detector \cite{alafif2017}. The datasets are publicly available at \url {http://discovery.cs.wayne.edu/lab_website/lsdl/} for research and non-commercial use only. The remainder of this paper is organized as follows. In {\bf Section II}, we review the existing the labeled face images datasets. 
\begin{figure}[h]
\begin{center}
\includegraphics[scale=.38]{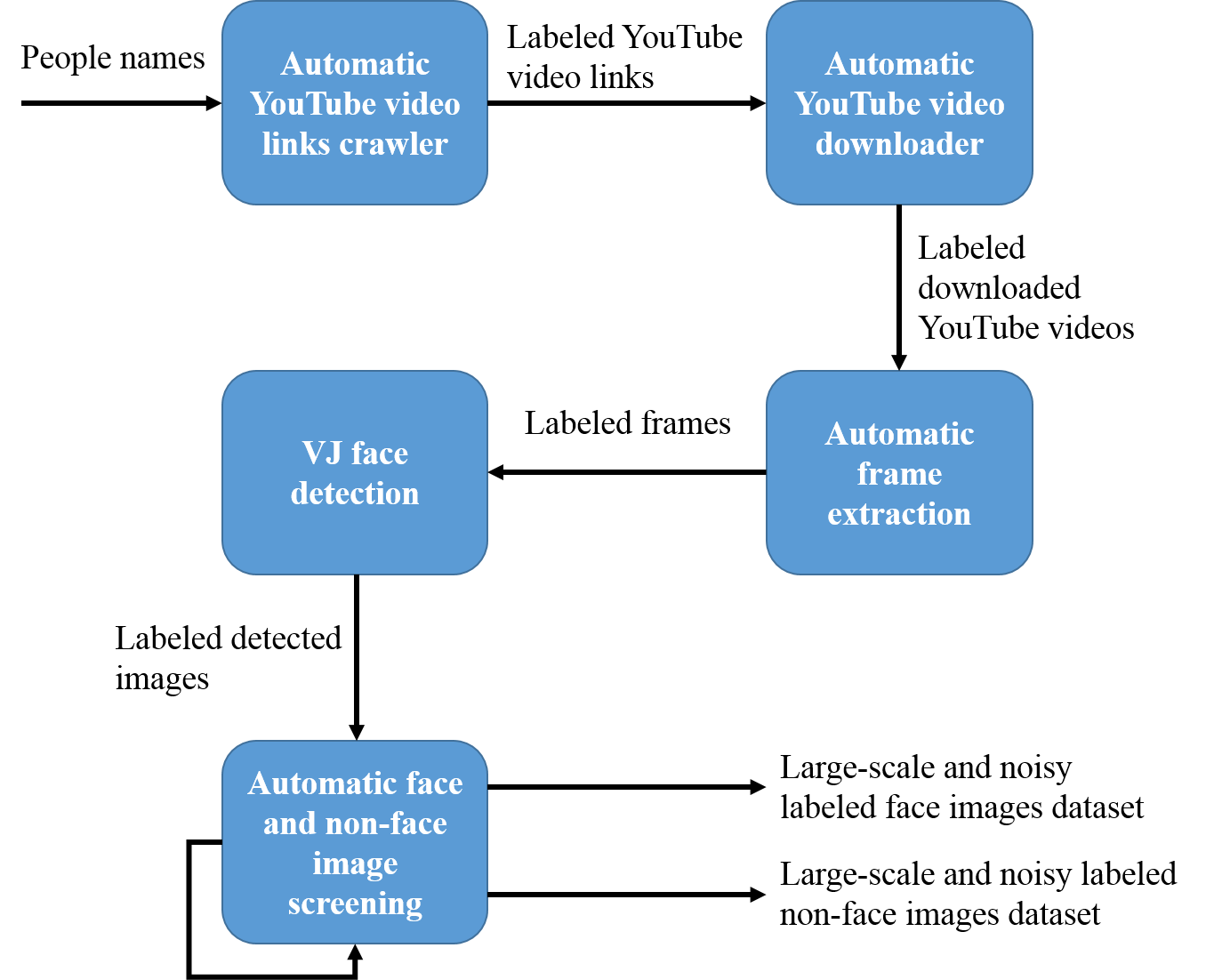}
\end{center}
   \caption{Phases for obtaining noisy LSLF and LSLNF datasets.}
\label{fig:sysarc}
\end{figure}
In {\bf Section III}, we first introduce the collection mechanism for obtaining our large-scale image datasets, Large-scale Labeled Face (LSLF) dataset and Large-scale Labeled Non-face (LSLNF) dataset. The LSLF dataset is compared with the existing labeled face image datasets. Then, we introduce our CrowedFaces and CrowedNonFaces image datasets. Finally, we conclude our work in {\bf Section IV}.

\section{Related Work}
Existing labeled face datasets have been available for different goals, but they tend to be in small-scales for face learning in both constrained and unconstrained environments. These datasets are briefly overviewed as follows: 

{\bf Labeled Faces in the Wild (LFW) Dataset} \cite{huang2007labeled}. The LFW dataset contains of 13,749 labeld face images for 5,749 individuals. The images were automatically labeled. This dataset was collected from web news articles. The face images in LFW dataset have variations in color and grayscale, near frontal pose, lighting, resolution, quality, age, gender, unbalanced race, accessory, partial occlusion, make up, and background. The dataset size is 179 MB. It is publicly available for downloading.

{\bf WebV-Cele Dataset} \cite{chen2014name}. The WebV-Cele dataset contains of 649,001 face images for 2,427 individuals. Only 42,118 face images are manually labeled. The face images were collected from YouTube videos. The images have variations in color and grayscale, quality, resolution, pose, illumination, background, human face, facial expression, partial occlusion, make up, gender, and race. The dataset is available upon request.

{\bf CAS-PEAL Dataset} \cite{gao2008cas}. The CAS-PEAL dataset contains of 99,594 manually labeled face images for 1,040 individuals. The face images were collected from a studio in constrained environment. The images have variations in color and grayscale , expression, lighting, pose, and accessory. This dataset only contains Chinese faces. The dataset size is nearly 26.6 GB. The dataset is available upon request.

{\bf Face Recognition Grand Challenge (FRGC) Dataset} \cite{phillips2005overview}. The FRGC dataset contains of nearly 50,000 manually labeled face images for 466 individuals. The face images were collected from a studio in constrained environment. The images have variations in color, lighting, expression, background, unbalanced race, 3D scan, and image sequence. The dataset size is 3.1 MB. The dataset is available upon request.

{\bf Multi-PIE Dataset} \cite{gross2010multi}. The Multi-PIE dataset contains of 755,370 manually labeled face images for 337 individuals. The face images were collected from a studio in constrained environment. The images have variations in color, resolution, pose, illumination, and expression. The dataset size is 308 MB. The dataset is commercial.

{\bf FERET Dataset} \cite{phillips2000feret}. The FERET dataset contains of 14,126 manually labeled face images for 1,199 individuals. The face images were collected from a studio in constrained environment. The images have variations in color, pose, and illumination. The dataset is available upon request.

{\bf Extended Yale B Dataset} \cite{georghiades2001few}. The Extended Yale B dataset contains of 16,128 face images for 28 individuals. The face images were collected from a studio in constrained environment. The images have variations in grayscale, pose, and illumination. The dataset is available upon request.

\section{Our Datasets}
In this section, we first describe the methodology for obtaining our LSLF dataset and noisy LSLNF dataset from the wild. Second, we introduce our CrowedFaces and CrowedNonFaces datasets. The datasets are briefly explained as follows:

\subsection{LSLF and LSLNF Datasets}
\begin{figure}[h]
\begin{center}
\includegraphics[scale=.75]{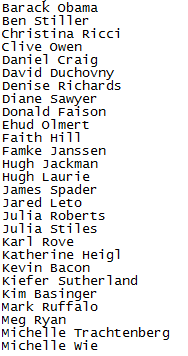}
\end{center}
   \caption{Sample from individual names.}
\label{fig:Names}
\end{figure}

We first used Wikipedia and other online resources to collect 11,690 popular names for individuals from many countries all over the world. The individual names include celebrities in many categories such as politics, sports, journalism, movies, arts, and educations. Most individual names consist in our datasets are associated with the individual’s first name then last name or vice versa while the rest of the names are famous nick names or popular single names. A sample from individual names is shown in Figure~\ref{fig:Names}.

We employ a systematic procedure consisting of five phases to obtain noisy LSLF and LSLNF image datasets as shown in Figure~\ref{fig:sysarc}. The five phases consist of automatic YouTube video links crawler, automatic YouTube video downloader, automatic framing extraction, Viola and Jones (VJ) face detection \cite{viola2001rapid}, and automatic face and non-face screening. The phases are briefly explained as follows:

\subsubsection{Automatic YouTube video links crawler}
\begin{figure}[h]
\begin{center}
\includegraphics[scale=.75]{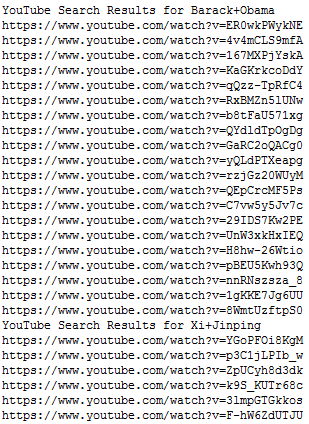}
\end{center}
   \caption{Sample from YouTube video links.}
\label{fig:links}
\end{figure}

In this phase, we developed a YouTube video links crawler to read the collected individual names and automatically retrieve and collect YouTube video links for each individual. The individuals’ YouTube video links still can be retrieved even If the ordering of the names is different. The links are associated and written for each labeled individual name in a text file. A sample from YouTube video links is shown in Figure~\ref{fig:links}.

\subsubsection{Automatic YouTube video downloader}
\begin{figure}[h]
\begin{center}
\includegraphics[scale=.65]{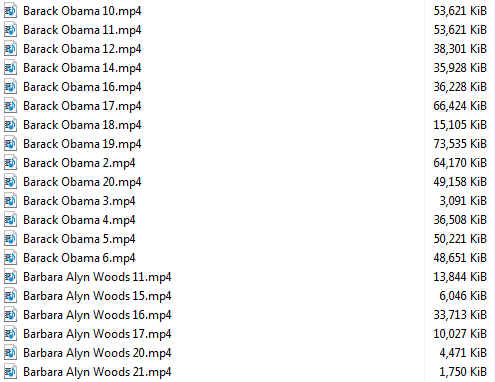}
\end{center}
   \caption{Sample from YouTube videos.}
\label{fig:videos}
\end{figure}

After retrieving YouTube video links using our automatic video links crawler, we developed a YouTube video downloader to automatically download the YouTube video links belonging to each individual. The labeled videos were stored to corresponding individual's labeled name. The total number of downloaded labeled videos is 129,435 videos with a total size of 2.96 TB.  All the labeled videos were stored in mp4 format for 11,690 individuals where individuals have a minimum of 1 video and a maximum of 68 videos. Our labeled YouTube video dataset has a larger number of labeled videos per person and a larger number of people among other existing labeled video datasets such as YouTube Faces \cite {wolf2011face} and WebV-Cele \cite {chen2014name}. A sample from YouTube videos is shown in Figure~\ref{fig:videos}.

\subsubsection{Automatic frame extraction}
\begin{figure}[h]
\begin{center}
\includegraphics[scale=.48]{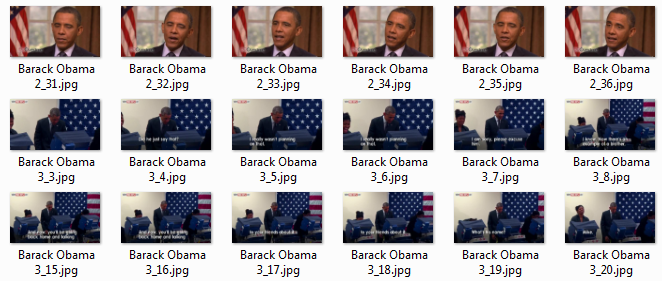}
\end{center}
   \caption{Sample frames from the frame dataset.}
\label{fig:Frame-Sample}
\end{figure}

After downloading the labeled YouTube videos, we randomly selected and extracted a number of frames per video. Following this procedure, we obtained 5,033,177 frames with a total size of 178 GB. All the frames were stored in JPEG format and labeled for 11,690 individuals where individuals have a minimum of 34 frames and a maximum of 2,716 frames. The average number of frames per individual is 430. A sample from the extracted frames is shown in Figure~\ref{fig:Frame-Sample}.

\subsubsection{VJ face detection}
\begin{figure}[h]
\begin{center}
\includegraphics[scale=.49]{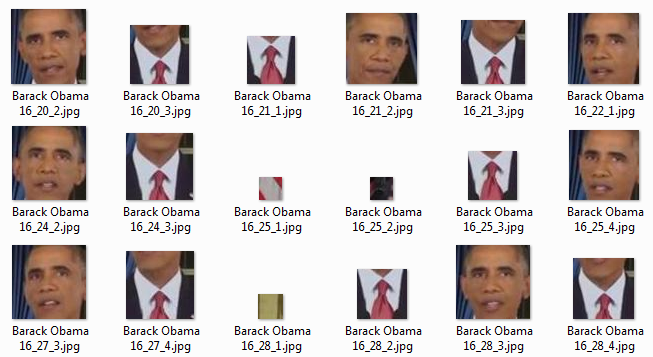}
\end{center}
   \caption{Sample images from the results of VJ face detection.}
\label{fig:VJfaceDetection}
\end{figure}

After extracting and storing the labeled frames, we automatically applied the VJ face detector to all labeled frames to detect faces. By applying VJ face detection, many face and non-face examples were detected due to a drawback of the VJ face detector that results in many false positives. Each detected example was automatically labeled and associated with the same name as used for the individuals. The number of labeled images is 9,750,456 with a total size of 14.4 GB. All the images were stored in JPEG format for 11,690 individuals where individuals have a minimum of 3 images and a maximum of 6,697 images. The average number of detected images per individual is 834. A sample of images from the results of VJ face detection is shown in Figure~\ref{fig:VJfaceDetection}.

\newcolumntype{C}{>{\centering\arraybackslash}X} 

\begin{table*}[htp]
\centering
\small
{\begin{tabular}{|C p{0.09\textwidth}| C p{0.09\textwidth}| p{0.1\textwidth}|p{0.07\textwidth}|p{0.09\textwidth}|p{0.1\textwidth}|p{0.2\textwidth}| C p{0.08\textwidth}|}
\hline
\rule{0pt}{2ex}
  \textbf{Face image dataset} & \textbf{\# of people} & \textbf{\# of face images} & \textbf{dataset size} & \textbf{Availability} & \textbf{Acquired} & \textbf{Image variations} & \textbf{Reference}
  \\ 
  \hline 
  \rule{0pt}{2ex}
 LFW & 5,749 & 13,233 automatically labeled & 179 MB & available &  web news articles & color and grayscale, most are near frontal poses, lightings, resolutions, quality, ages, gender, unbalanced races, accessories, partial occlusions, make up, backgrounds & \cite{huang2007labeled}  \\ 
\hline
\rule{0pt}{2ex}
  WebV-Cele & 2,427 & 649,001 but only
  42,118 are manually labeled for 144 people
   & unknown & on request &  YouTube videos & color and grayscale, quality, resolutions, poses, illuminations, backgrounds, human faces, facial expressions, partial occlusions, make up, gender, races, accessories, tattoos, hats, glasses, sunglasses, hands, scarves, microphones, or other objects or persons & \cite{chen2014name}  \\ 
  \hline
  \rule{0pt}{2ex}
  CAS-PEAL & 1,040 & 99,594 manually labeled & $>$26.6 GB & on request &  studio & color and grayscale , expressions, lightings, poses, accessories, Chinese & \cite{gao2008cas}  \\ 
  \hline
  \rule{0pt}{2ex}
  Face Recognition Grand Challenge & $>$466 & $>$50,000 manually labeled & 3.1 MB & on request &  studio & color, lightings, expressions, backgrounds, unbalanced races, 3D scans, image sequences & \cite{phillips2005overview}  \\ 
  \hline
  \rule{0pt}{2ex}
  Multi-PIE & 337 & 755,370 manually labeled & 308 GB & commercial & studio & color, resolutions, poses, illuminations, expressions & \cite{gross2010multi}  \\ 
  \hline
  \rule{0pt}{2ex}
  FERET & 1,199 & 14,126 manually labeled & unknown & on request & studio & color, poses, expressions & \cite{phillips2000feret}  \\ 
  \hline
  \rule{0pt}{2ex}
  Extended Yale B & 28 & 16,128 & 4.68 GB & available & studio & grayscale, poses, illuminations & \cite{georghiades2001few}  \\ 
  \hline
  \rule{0pt}{2ex}
  {\bf Ours LSLF} & 11,459 & 1,195,976 automatically labeled & 5.36 GB & available & YouTube videos & color and grayscale, quality, resolutions, multi-view, illuminations, backgrounds, illusions, human faces, cartoon faces, facial expressions, make up, tattoos, gender, ages, races, and partial occlusions by accessories, hats, glasses, sunglasses, hands, hair, beards, scarves, microphones, and other objects or persons & \textbf{-}  \\ 
   \hline
\end{tabular}}
\newline \newline
\caption{A comparison of existing face datasets and our LSLF dataset.} \label{table1}

\end{table*} 

\subsubsection{Automatic face and non-face image screening}

In this phase, we automatically separated out face and non-face examples as much as possible from the results of the VJ face detection using several automatic screenings. For automatic image screenings, several trained classification models for facial parts were applied to the results of the VJ face detection phase.

\begin{figure}[h]
\begin{center}
\includegraphics[scale=.87]{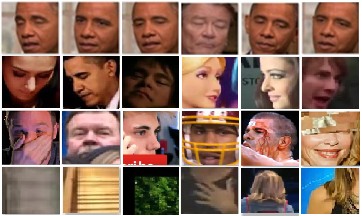}
\end{center}
   \caption{Sample images from the LSLF and noisy LSLNF datasets. The first three rows present face images from the LSLF dataset. Frontal and near frontal faces are presented in the first row. Multi-view faces are presented in the second row. Partially occluded faces are presented in the third row. The last row presents non-face images from the noisy LSLNF dataset.}
\label{fig:imagesample2}
\end{figure}

After completing this screening phase, we obtained the LSLF and LSLNF datasets with slight noise. Here, the noise referred to is the false positives resulted after classification. To be specific, LSLF contains about 1.7\% non-faces while LSLNF contains about 10\% faces. After completing the screening phase, individuals with 0 images are automatically removed from both datasets. Therefore, our LSLF dataset consists of 1,217,185 labeled face images for 11,478 individuals with about 1.7\% noise. These images are stored in JPEG format with a total size of 3.42 GB. Individuals have a minimum of 1 face image and a maximum of 1,177 face images. The average number of face images per individual is 106. On the other hand, our noisy LSLNF dataset consists of 3,468,430 labeled none-face images for 11,682 individuals with about 10\% noise. These images are stored in JPEG format with a total size of 3.16 GB. Individuals have a minimum of 1 non-face image and a maximum of 2,282 non-face images. The average number of non-face images per individual is 296.

After obtaining the noisy LSLF dataset, we manually removed the noise from it. Therefore, our LSLF dataset consists of 1,195,976 labeled face images for 11,459 individuals. These images are stored in JPEG format with a total size of 5.36 GB. Individuals have a minimum of 1 face image and a maximum of 1,157 face images. The average number of face images per individual is 104. Each image is automatically named as (PersonName\_VideoNumber\_FrameNumber\_ImageNuumber) and stored in the related individual folder. Image samples from the LSLF and noisy LSLNF datasets are shown in Figure~\ref{fig:imagesample2}.

Our LSLF dataset consists of multi-view faces. Many of these faces have frontal and near frontal poses. These faces have large image variations in color and grayscale, image quality, image resolution, image illumination, image background, image illusion, human face, cartoon face, facial expression, light and severe partial facial occlusion, make up, gender, age, and race. Also, our LSLF dataset has a broad distribution of races from different parts of the world. Many of our face images are partially occluded with accessories such as tattoos, hats, glasses, sunglasses, hands, hair, beards, scarves, microphones, or other objects or persons. These factors essentially make our dataset great for large scale face learning and face recognition tasks. To the best of our knowledge, our LSLF dataset is the largest labeled face image dataset in the literature in terms of the number of labeled images and the number of individuals compared to other existing face image datasets \cite{huang2007labeled, chen2014name, gao2008cas, phillips2005overview, gross2010multi, phillips2000feret, georghiades2001few}. A brief comparison is made in Table~\ref{table1}. 

\subsection{CrowdFaces And CrowdNonFaces Datasets}
\begin{figure}[h]
\begin{center}
\includegraphics[scale=.96]{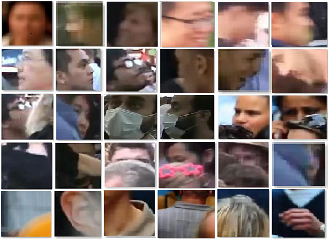}
\end{center}
   \caption{Sample images from the CrowdFaces and CrowdNonFaces datasets. The first four rows are a sample of faces from the CrowdFaces dataset. Multi-view blurred faces are presented in the first row. Multi-view non-blurred faces are presented in the second row. Multi-view light partially occluded faces are presented in the third row. Multi-view severe partially occluded faces are presented in the fourth row. The last row is a sample of non-faces from the CrowdNonFaces dataset.}
\label{fig:POF2}
\end{figure}

We introduce our two other datasets extracted from the wild, CrowdFaces and CrowdNonFaces datasets. The objective of these datasets is to obtain multi-view blurred and non-blurred faces, multi-view partially occluded faces, and non-faces to be used for the training. For obtaining these datasets, we manually selected and downloaded thirty YouTube videos that include crowd scenes in streets, sport games, religious gatherings, parties, street fights, and courts. Partially occluded faces are occluded by overlapping objects such as faces, hands, bodies, hats, masks, hair, etc. An overlapping sliding window is applied to these scenes at all locations with different scales to extract non-faces, multi-view blurred and non-blurred faces, and multi-view partially occluded faces. The sliding window starts at size 16 x 16 and increases by 1.5 scaling factor until the window size is no larger than the frame. We manually curated 10,049 faces and 31,662 non-faces at different scales from these sub-windows. Our partially occluded face images include light and severe partially occluded faces. A sample of these images is shown in Figure~\ref{fig:POF2}.

\section{Conclusion}
In this paper, we first introduce our large-scale image datasets, LSLF and noisy LSLNF. Our LSLF dataset consists of a large number of unconstrained multi-view and partially occluded faces. The faces have many variations in color and grayscale, quality, resolution, illumination, background, illusion, human face, cartoon face, facial expression, light and severe partial facial occlusion, make up, gender, age, and race. Many of the face images are partially occluded with accessories such as tattoos, hats, glasses, sunglasses, hands, hair, beards, scarves, microphones, or other objects or persons. The LSLF dataset is currently the largest labeled face image dataset in the literature in terms of the number of labeled images and the number of individuals compared to other existing face image datasets. Second, we introduced our CrowedFaces and CrowedNonFaces image datasets. The CrowedFaces and CrowedNonFaces datasets include faces and non-faces images from crowed scenes. These datasets essentially aim for researchers to provide a large number of training examples with many variations for large scale face learning and face recognition tasks.



%



\bibliographystyle{ieeetran}
\bibliography{ref1}

\end{document}